%% file: mesh21.tex
\def\year{2021}\relax
\documentclass[letterpaper]{article} 
\usepackage{aaai21}  
\usepackage{times}  
\usepackage{helvet} 
\usepackage{courier}  
\usepackage[hyphens]{url}  
\usepackage{graphicx} 
\usepackage{natbib}  
\usepackage{caption} 
\input{math_commands.tex}

\usepackage{multirow}
\usepackage[utf8]{inputenc} 
\usepackage[T1]{fontenc}    
\usepackage{booktabs} 

\title{Learning Local Neighboring Structure for Robust 3D Shape Representation}
\author {
        Zhongpai Gao\textsuperscript{\rm1},
        Junchi Yan\textsuperscript{\rm1},
        Guangtao Zhai\textsuperscript{\rm1,2}\thanks{Corresponding author.},
        Juyong Zhang\textsuperscript{\rm3},
        Yiyan Yang\textsuperscript{\rm1},
        Xiaokang Yang\textsuperscript{\rm1}\\
}
\affiliations {
    \textsuperscript{\rm 1} MoE Key Lab of Artificial Intelligence, AI Institute, Shanghai Jiao Tong University \\
    \textsuperscript{\rm 2} Institute of Image Communication and Network Engineering, Shanghai Jiao Tong University \\
    \textsuperscript{\rm 3} School of Mathematical Sciences, University of Science and Technology of China \\
    \{gaozhongpai, yanjunchi, zhaiguangtao\}@sjtu.edu.cn, juyong@ustc.edu.cn, \{jamson, xkyang\}@sjtu.edu.cn
}
\begin{document}

\maketitle

\begin{abstract}
Mesh is a powerful data structure for 3D shapes. Representation learning for 3D meshes is important in many computer vision and graphics applications. The recent success of convolutional neural networks (CNNs) for structured data (e.g., images) suggests the value of adapting insight from CNN for 3D shapes. However, 3D shape data are irregular since each node's neighbors are unordered. Various graph neural networks for 3D shapes have been developed with isotropic filters or predefined local coordinate systems to overcome the node inconsistency on graphs. However, isotropic filters or predefined local coordinate systems limit the representation power. In this paper, we propose a local structure-aware anisotropic convolutional operation (LSA-Conv) that learns adaptive weighting matrices for each node according to the local neighboring structure and performs shared anisotropic filters. In fact, the learnable weighting matrix is similar to the attention matrix in the random synthesizer -- a new Transformer model for natural language processing (NLP). Comprehensive experiments demonstrate that our model produces significant improvement in 3D shape reconstruction compared to state-of-the-art methods.
\end{abstract}

\section{Introduction}

Representation learning for 3D meshes is crucial for many 3D tasks, e.g., reconstruction \cite{Genova_2018_CVPR, Tran_2018_CVPR, gao2020semi}, shape correspondence \cite{Groueix_2018_ECCV}, shape synthesis and modeling \cite{6654137, Xie_2018_CVPR}, face recognition \cite{Liu_2018_CVPR} and shape segmentation \cite{djonlic2017segmentation, 10.1145/1833349.1778839}, and graphics applications such as virtual avatar \cite{Cao2014}. Inspired by the great success of convolutional neural networks (CNN) in the fields of natural language processing (1-D), image classification (2-D), and radiographic image analysis (3-D) where underlying data are Euclidean structured, deep neural networks on 3D meshes have recently driven significant interests. Directly applying CNN on 3D meshes is a challenge since they are non-Euclidean structured and are usually represented as graphs in which the number and orientation of each node's neighbors vary from one to another (node inconsistency). An effective definition of convolutional operation analogous to that on Euclidean structured data is important for 3D shape representation learning.

Recently, many graph convolutional networks have been developed to handle irregular graph data and achieved promising results. \citet{NIPS2016_6081} designed fast localized convolutional filters on graphs using Chebyshev expansion based on spectral graph theory, called \emph{ChebNet}. The spectral filters are isotropic to overcome the irregularity of graph data. ChebNet is an efficient generalization of CNNs to graphs. \citet{Ranjan_2018_ECCV} built convolutional mesh autoencoder (COMA) for 3D meshes with fixed topology upon ChebNet and introduced mesh sampling operations that enable a hierarchical representation to capture non-linear variations of human faces. However, compared to CNN, isotropic filters used in ChebNet limit the representation power.

In order to introduce anisotropic filters on graph convolutions, \citet{Bouritsas_2019_ICCV} formulated a spiral convolution operator (SpiralNet) that defines a explicit order of the neighbors via a spiral scan for each vertex on 3D meshes with fixed topology. However, serializing the local neighbors of vertices by following a spiral cannot resolve the inconsistency between different nodes. Furthermore, this method requires manually assigning a starting point to determine the order of neighbors, which is difficult to make the local coordinate system consistent across meshes. The selection of the starting point may affect the performance of the spiral convolution operator. Explicitly defining the order of neighbors cannot efficiently exploit the irregular structure of graphs.

PointCNN \cite{NIPS2018_7362} proposed to learn an $\mathcal{X}$-transformation from the input points to weight and permute each point's neighbors into a latent and potentially canonical order. KPConv \cite{Thomas_2019_ICCV} presented a convolutional operation that weights each point's neighbors depending on the Euclidean distances to a set of predefined or deformable kernel points. Subsequently, these methods apply anisotropic filters on the resampled neighbors to extract features for point clouds. However, different from 3D meshes, point clouds do not have a fixed topology and the neighbors are obtained through a $K$-nearest neighbors (KNN) algorithm. The methods designed for point clouds do not consider the unique characteristics of 3D meshes, which limits the representation power for 3D meshes.

This paper proposes a novel \textbf{l}ocal \textbf{s}tructure-aware \textbf{a}nisotropic convolutional operation (LSA-Conv) for 3D meshes. Consider that 3D meshes are irragular and share the same topology of a template, e.g., 3D morpable models (3DMM) \cite{5279762, 6654137, SMPL2015, MANO:SIGGRAPHASIA:2017}, we directly learn a weighting matrix for each vertex of the template to soft-permute the vertex's neighbors. The weighting matrices are trained along with the whole network. The idea of learnable weighting matrix is by analogy with the random synthesizer \cite{tay2020synthesizer}, which is a new Transformer model for natural language processing (NLP). Then similar to CNNs, we apply shared anisotropic filters on the resampled neighbors to extract local features on 3D meshes. LSA-Conv is designed to adapt each vertex's local structure without explicitly defining the order or any local pseudo-coordinate systems for each vertex.

LSA-Conv is easy to implement and integrate into existing deep learning models to improve their performance. In line with COMA \cite{Ranjan_2018_ECCV}, SpiralNet \cite{Bouritsas_2019_ICCV}, and SpiralNet++ \cite{gong2019spiralnet++} that are all designed for meshes with fixed topologies, we evaluate our approach on the reconstruction task which has been a fundamental testbed for further applications. Fixed topology is a common and practical setting for meshes in face, hand, and body related applications \cite{gao2020semi, Jiang_2019_CVPR}. Note that, meshes with arbitrary topologies can directly be handled by point-cloud based methods \cite{Verma_2018_CVPR, NIPS2018_7362}. The proposed method is complementary to these point-cloud works and makes its unique value for fixed-topology based applications. We use LSA-Conv to build convolutional mesh autoencoder and achieve state-of-the-art performance on two 3D shape datasets: human faces (COMA \cite{Ranjan_2018_ECCV}) and human bodies (DFAUST \cite{Bogo_2017_CVPR}). Comprehensive evaluation experiments show that the proposed method significantly outperforms existing models.

The contributions of this paper are summarized in below:

1) Taking advantage of the readily available fixed-topology information of mesh data, we propose a \textbf{l}ocal \textbf{s}tructure-aware \textbf{a}nisotropic convolutional operation (LSA-Conv) for representation learning from 3D meshes. LSA-Conv learns a weighting matrix for each vertex to soft-permute its neighbors based on the local neighboring structure as derived from the object-level topology. The learnable weighting matrix is similar to the attention matrix in random synthesizer \cite{tay2020synthesizer} for NLP. To our knowledge, this is the first work for learning local arrangement by utilizing the fixed-topology, in contrast to the non-adaptive methods either using isotropic filters~\cite{Ranjan_2018_ECCV} or predefined local coordinates~\cite{Bouritsas_2019_ICCV}.

2) Unlike \cite{Bouritsas_2019_ICCV} that needs to define the neighboring order explicitly, LSA-Conv needs no pre/post processing steps. Thus, LSA-Conv is orthogonal to other techniques for 3D meshes and can be readily integrated into existing pipelines~\cite{Ranjan_2018_ECCV, gao2020semi} for 3D shape processing, by replacing the conv layer.

3) Extensive experiments show that our model significantly outperforms state-of-the-art methods for 3D shape generations. Two applications demonstrates the effectiveness of the proposed method. The source code will be made public available.

\section{Related Work}
\textbf{Linear 3D morphable models} 3D morpable models (3DMM) are statistical models of 3D shapes, such as human faces, bodies, hands, etc., and are constructed by performing some form of dimensionality reduction on a training set that each mesh is in dense correspondence with each other (i.e., fixed topology). 3DMMs are powerful priors on 3D shape reconstruction or generation. \citet{blanz1999morphable} proposed the first linear parametric 3DMM using principal component analysis (PCA) to model the shape and texture of 3D faces. The widely used 3DMM for faces \cite{7298679} was built by merging Basel Face Model (BFM) \cite{5279762} with 200 subjects in neutral expressions and FaceWarehouse \cite{6654137} with 150 subjects in 20 different expressions. Skinned multi-person linear model (SMPL) \cite{SMPL2015} is the most well known body model as learned through PCA and represents a wide variety of body shapes in natural human poses. MANO \cite{MANO:SIGGRAPHASIA:2017} is a hand model learned from around 1000 high-resolution 3D scans of human hands in a wide variety of hand poses. Those PCA-based models are commonly used for 3D faces, bodies, and hands reconstruction. In this paper, we introduce a non-linear 3DMM for 3D shapes with much higher representation power.

\textbf{Graph neural networks} The popularity of extending deep learning approaches for graph data has been rapidly growing in recent years. Convolutional graph neural networks fall into two categories: spectral-based and spatial-based. Spectral-based approaches define convolutional operation based on graph signal processing. Spectral CNN \cite{BrunaZSL13} generalizes convolution to graphs via Laplacian eigenvectors. ChebNet \cite{NIPS2016_6081} and GCN \cite{kipf2017semi} reduce the computation complexity of eigen-decomposition by using fast localized convolutional filters. AGCN \cite{li2018adaptive} learns hidden structural relations unspecified by the graph adjacency matrix.
Spatial-based approaches define graph convolutions based on a node's spatial relations. GraphSage \cite{NIPS2017_6703} samples a fixed number of neighbors and aggregates neighboring features for each node. GAT \cite{velickovic2018graph} adopts attention mechanisms to learn the relative weights between two connected nodes. MoNet \cite{Monti_2017_CVPR} introduces node pseudo-coordinates to determine the relative position between a node and its neighbors and assigns different weights to the neighbors. FeaStNet \cite{Verma_2018_CVPR} proposes a graph-convolution operator that learns a weighting matrix dynamically computed from features, which is similar to PointCNN \cite{NIPS2018_7362}.

\textbf{Point neural networks} Several point neural networks are related to our work. PointCNN \cite{NIPS2018_7362} presents a method to learn an $\mathcal{X}$-transformation as a function of input points. RandLA-Net \cite{hu2019randla} uses an attention mechanism to learn local features instead of applying filters for convolutional operation. KPConv \cite{Thomas_2019_ICCV} proposes a convolution that takes radius neighbors as input and processes them with weights spatially located by a set of kernel points. Instead of calculating a weighting matrix as a function of inputs in FeaStNet \cite{Verma_2018_CVPR}, PointCNN \cite{NIPS2018_7362}, and KPConv \cite{Thomas_2019_ICCV}, our LSA-Conv directly learns a weighting matrix for each vertex thanks to the fixed topology of meshes which otherwise is unavailable in cloud data.


\section{Approach}

\begin{figure}[tb!]
    \centering
    \includegraphics[width=0.48\textwidth]{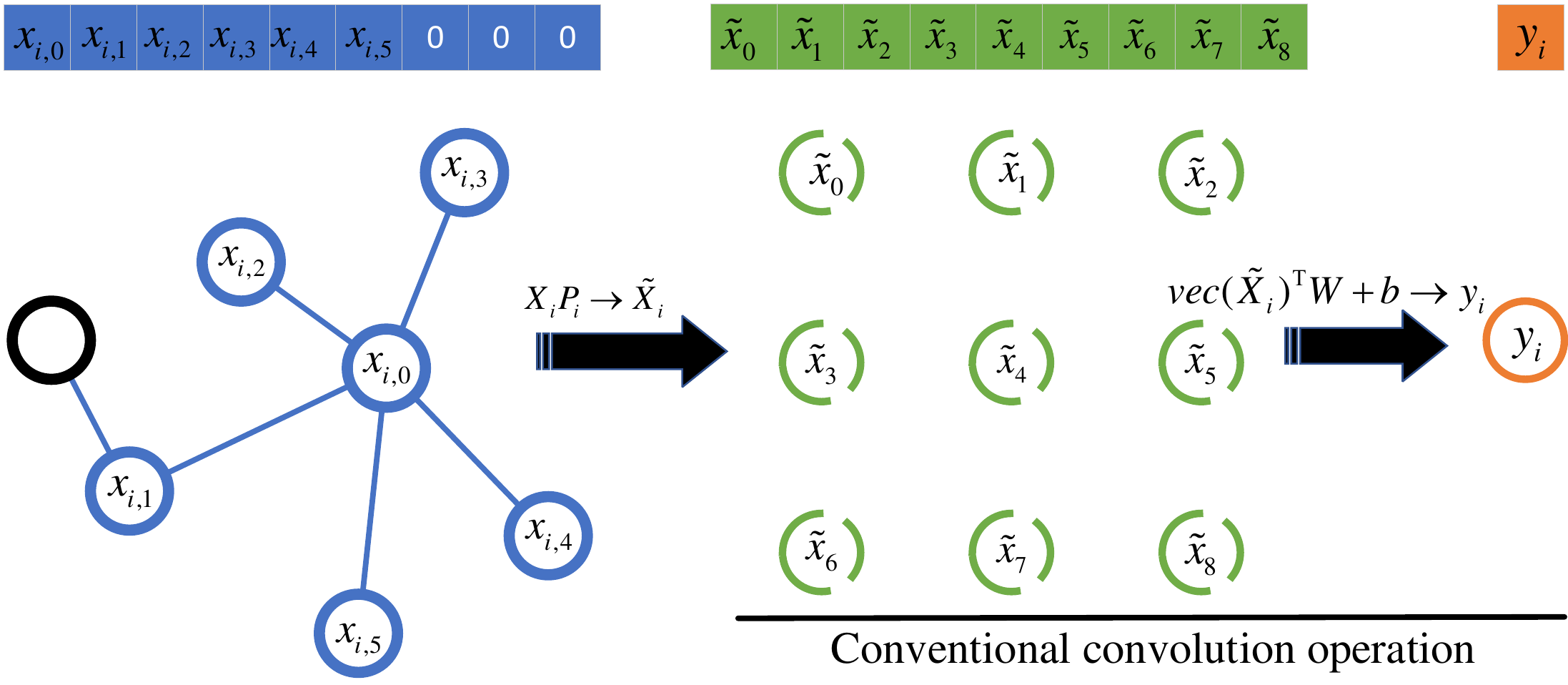}
\caption{Local structure-aware anisotropic convolutional operation (LSA-Conv). $\{\vx_{i,0}, \vx_{i, 1}, \ldots,$ $\vx_{i, 5}\} \subseteq \bm{\mathcal{N}}_i$ are the one-ring neighborhood of $\vx_i$ (including itself, i.e., $\vx_{i,0}$). We construct $\mX_i = \{\vx_{i,0}, \vx_{i, 1}, $ $ \ldots, \vx_{i, K-1}\} \in \R^{D_{in} \times K}$ where $K$ is a predefined neighbor size. LSA-Conv contains: i) a learnable weighting matrix, $\mP_i\in\R^{K \times K}$ is used to soft-permute the node's neighbors; ii) a conventional convolution operation with anisotropic filters, $\mW\in\R^{(D_{in}\cdot K)\times D_{out}}$, is performed with bias $\vb\in\R^{D_{out}}$.}
\label{fig:LSA-gcn}
\end{figure}

Aiming at entailing the GCNs with the anisotropic filtering ability like CNNs to improve its expressiveness, we propose a local structure-aware anisotropic convolutional operation (LSA-Conv) on graphs. Instead of directly applying anisotropic filters on each node's neighbors, we soft-permute each node's neighbors using a weighting matrix that is trained along with the deep neural networks.

\subsection{Local structure-aware anisotropic convolution}
\label{sec:LSA_conv}
Consider a 3D shape that is described as a mesh $M = (\bm{\mathcal{V}}, \bm{\mathcal{E}})$, where $\bm{\mathcal{V}}=\{1, \ldots, N\}$ is a set of vertices and $\bm{\mathcal{E}}\subseteq \bm{\mathcal{V}} \times \bm{\mathcal{V}}$ is a set of edges. A graph may have node attributes $\mX \in \R^{D \times N}$, where $D$ and $N$ represents the feature dimension and number of nodes, respectively. In the simplest setting of $D=3$, each node contains 3D coordinates $\vx_i=[x_i, y_i, z_i]^\top$ in the Euclidean space. The node attributes can also include additional coordinates such as color and vertex normal. In a deep neural network, the output of each layer is as the input for the subsequent layer. Thus, generally, $D$ represents the feature dimension of a given layer in the deep neural network.

We define LSA-Conv as follows. For each node, $\bm{\mathcal{N}}_i$ is a set of the one-ring neighborhood of $\vx_i$ (including itself, also as $\vx_{i,0}$), where $\vx_j \in \bm{\mathcal{N}}_i$ and $(\vx_{i,0}, \vx_{i,j}) \in \bm{\mathcal{E}}$. We denote $\bm{\mathcal{N}}_i = \{\vx_{i,0}, \vx_{i, 1}, \ldots, \vx_{i, |\bm{\mathcal{N}}_i|-1}\}$, where $|\bm{\mathcal{N}}_i|$ is the number of node's neighbors and it varies from one node to another in a graph. In order to apply a shared anisotropic filter on each node, we define a constant neighbor size $K$, which corresponds to the kernel size in conventional convolution operation. For each node, we construct $\mX_i = \{\vx_{i,0}, \vx_{i, 1}, \ldots, \vx_{i, K-1}\} \in \R^{D_{in} \times K}$, where the first $K$ neighbors are selected if $K$ is smaller than or equal to $|\bm{\mathcal{N}}_i|$; otherwise zero-padding is applied, as shown in Figure \ref{fig:LSA-gcn}. Note that, we put the node itself, $\vx_i$, in the first place and the order of other neighbors $\{\vx_{i, 1}, \ldots, \vx_{i, K-1}\}$ in $\mX_i$ is random and not specified.

Since the order and orientation of neighbors for each node vary from one to another, directly applying an anisotropic filter on unordered neighbors diminishes the representation power. While training, the anisotropic filter may struggle to adapt to the large variability of the unordered coordinate systems and tend to become a isotropic (rotational invariance) filter. In this paper, we introduce an adaptive weighting matrix to soft-permute each node's neighbors, denoted as $\mP_i\in \R^{K\times K}$. The resampled convolutional neighbors of each node can be obtained by
\begin{align}\label{eq:permutation}
\tilde{\mX}_i = \mX_i \mP_i,
\end{align}
where $\tilde{\mX}_i\in \R^{D_{in} \times K}$ and $\mP_i$ is a trainable parameter to be adaptive according to the geometric structure of the node's neighbors.

\begin{figure*}[tb!]
    \centering
    \includegraphics[width=1.0\textwidth]{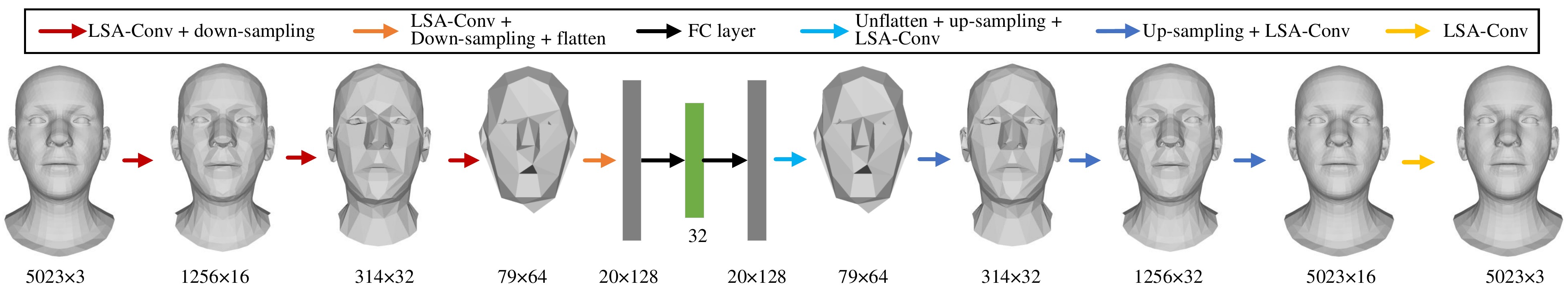}
\caption{Architecture of our LSA-Conv based 3D morphable models (LSA-3DMM).}
\label{fig:architecture}
\end{figure*}

The learned weighting matrix for each node can soft-permute the node'neighbors into an implicit canonical order such that we are able to apply a shared anisotropic filter on each node of a graph. This operation is the same as the conventional convolution and can be expressed as
\begin{align}\label{eq:conv}
\vy_i = \text{vec}(\tilde{\mX}_i)^\top \mW + \vb,
\end{align}
where $\mW\in\R^{(D_{in}\cdot K) \times D_{out}}$ includes $D_{out}$ anisotropic filters, $\vb\in\R^{D_{out}}$ is the bias, $\vy_i\in\R^{D_{out}}$ is the output feature node corresponding to the input node $\vx_i\in\R^{D_{in}}$, and $\text{vec}(\cdot)$ is a vectorization function which converts a matrix into a column vector. To introduce non-linearity, an activation function $f(\cdot)$ such as ELU \cite{clevert2015fast} is introduced on Eq. (\ref{eq:permutation}) and (\ref{eq:conv}). Thus, LSA-Conv is defined as
\begin{align} \label{eq:resample}
\vy_i = f\left(\text{vec}(f(\mX_i\mP_i))^\top\mW + \vb\right).
\end{align}
Note, for meshes with fixed topology, each node corresponds to a weighting matrix and all nodes of the whole graph share the same anisotropic filter for each output channel.

\subsection{LSA-Conv based 3D morphable models}
We propose a nonlinear 3D morphable model using our LSA-Conv as building blocks, called \emph{LSA-3DMM}. The basic architecture of LSA-3DMM is the same as COMA \cite{Ranjan_2018_ECCV} and Spiral \cite{Bouritsas_2019_ICCV} for a fair comparison, as shown in Figure \ref{fig:architecture}. The mesh sampling operations are adopted from \cite{Ranjan_2018_ECCV}.   We simply replace the convolutional operations (i.e., ChebNet or SpiralNet) with LSA-Conv. LSA-3DMM is a deep convolutional mesh autoencoder with hierarchical mesh representations and is able to capture nonlinear variations in 3D shapes at multiple scales within the model.

We denote as $FC(\cdot)$ a fully connected layer, $d$ the dimension of latent vector, $l$ the number of vertices after the last down-sampling layer, $PC(k, c)$ a LSA-Conv layer with neighbor size $k$ and number of filters $c$, $DS(p)$ and $US(p)$ are a down-sampling and a up-sampling layer by a factor of $p$, respectively. LSA-3DMM is listed as follows:
\begin{small}
\begin{align}
enc:&PC(9, 16)\!\rightarrow DS(4)\!\rightarrow PC(9, 32)\!\rightarrow DS(4)\!\rightarrow \nonumber\\
    &PC(9, 64)\!\rightarrow DS(4)\!\rightarrow PC(9, 128)\!\rightarrow DS(4)\!\rightarrow FC(d), \nonumber\\
dec:&FC(l*128)\!\rightarrow US(4)\!\rightarrow PC(9, 64)\!\rightarrow US(4)\!\rightarrow \nonumber\\
    &PC(9, 32)\!\rightarrow US(4)\!\rightarrow PC(9, 32)\!\rightarrow US(4)\!\rightarrow \nonumber\\
    &PC(9, 16)\!\rightarrow PC(9, 3).\nonumber
\end{align}
\end{small}The model's encoder effectively compresses a 3D shape into a low dimensional latent vector (e.g., $d=32$ in Figure \ref{fig:architecture}) and the decoder reconstructs the 3D shape from the latent vector. LSA-3DMM can be used in 3D shape recognition, reconstruction, and many other applications.

\section{Evaluation and Discussion}

In this section, we first evaluate the proposed model on two different 3D shape datasets by comparing to state-of-the-art approaches. Then, a parameter reduction method is proposed for LSA-Conv. At last, ablation tests are conducted to demonstrate the effectiveness of LSA-Conv.

\textbf{Datasets} We evaluate our model on two datasets: COMA \cite{Ranjan_2018_ECCV} and DFAUST \cite{Bogo_2017_CVPR}. COMA is a human facial dataset that consists of 12 classes of extreme expressions from 12 different subjects. The dataset contains 20,466 3D meshes that were registered to a common reference template with 5023 vertices. DFAUST is a human body dataset that collects over 40,000 real meshes, capturing 129 dynamic performances from 10 subjects. A mesh registration method that uses both 3D geometry and texture information to register all scans in a sequence to a common reference topology with 6890 vertices.
The same as in \cite{Ranjan_2018_ECCV}, we split both two datasets into training and test set with a ratio of 9:1 and randomly select 100 samples from the training set for validation. We perform standardization on all the 3D shape meshes by subtracting the mean shape and dividing with each vertex's standard deviation to improve the convergence speed of training.

\textbf{Training} We use Adam \cite{kingma2014adam} optimizer with learning rate 0.001 and reduce the learning rate with decay rate 0.99 in every epoch. The batch size is 32 and total epoch number is 300. We initialize the weighting matrices with identity matrix, $\mI\in\R^{K\times K}$, i.e., the network starts without a predefined order for the node's neighbors. Weight decay regularization is used for the network parameters except for the weighting matrices.

\subsection{Comparison to existing methods}

\begin{figure}[tb!]
    \centering
    \includegraphics[width=0.49\textwidth]{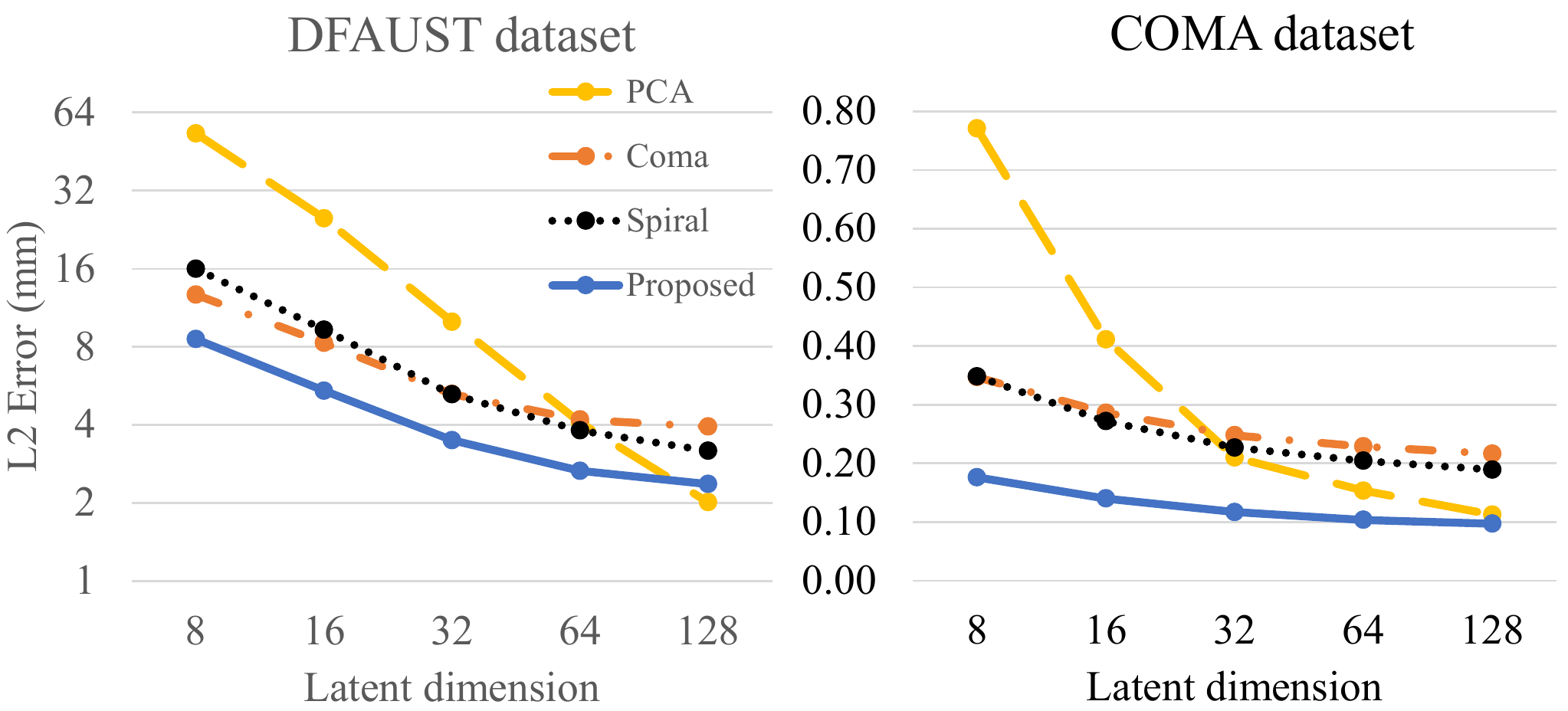}
\caption{Evaluation of LSA-3DMM against peer methods: PCA, COMA, and Spiral on test sets.}
\label{fig:evaluation}
\end{figure}

We compare three existing methods: PCA \cite{Blanz99}, COMA \cite{Ranjan_2018_ECCV}, and Spiral \cite{Bouritsas_2019_ICCV} on different dimensionalities of the latent space: 8, 16, 32, 64, and 128. The same architecture in Figure \ref{fig:architecture} is used in the methods of COMA, Spiral, and our LSA-3DMM for consistency and fair comparison. As shown in Figure \ref{fig:evaluation}, the proposed LSA-3DMM achieves the smallest reconstruction errors compared to COMA and Spiral on both COMA and DFAUST datasets by a large margin. LSA-3DMM consistently cuts the errors by around half for all the dimensionalities of latent space on both two datasets thanks to the proposed LSA-Conv operation. LSA-Conv significantly improves the expressive power on 3D shape representation learning compared to ChebNet and SpiralNet.

Compared to PCA, all the methods based on deep neural networks (DNN) have smaller reconstruction errors for a small latent size ($d < 32$). This is because PCA-based linear 3DMMs can only capture global features and DNN-based nonlinear 3DMMs are able to capture local features using convolutional operations. The reconstruction accuracy of LSA-3DMM is comparable to PCA even when the latent size is 128. Note that, small latent size is favorable. Smaller latent size makes each latent feature more semantically meaningful. Furthermore, directly applying PCA needs a large amount of memory for a large dataset. At last, in real applications, the model is trained on batch. PCA parameters cannot be updated with additional data.

\begin{figure*}[tb!]
    \centering
    \includegraphics[width=1\textwidth]{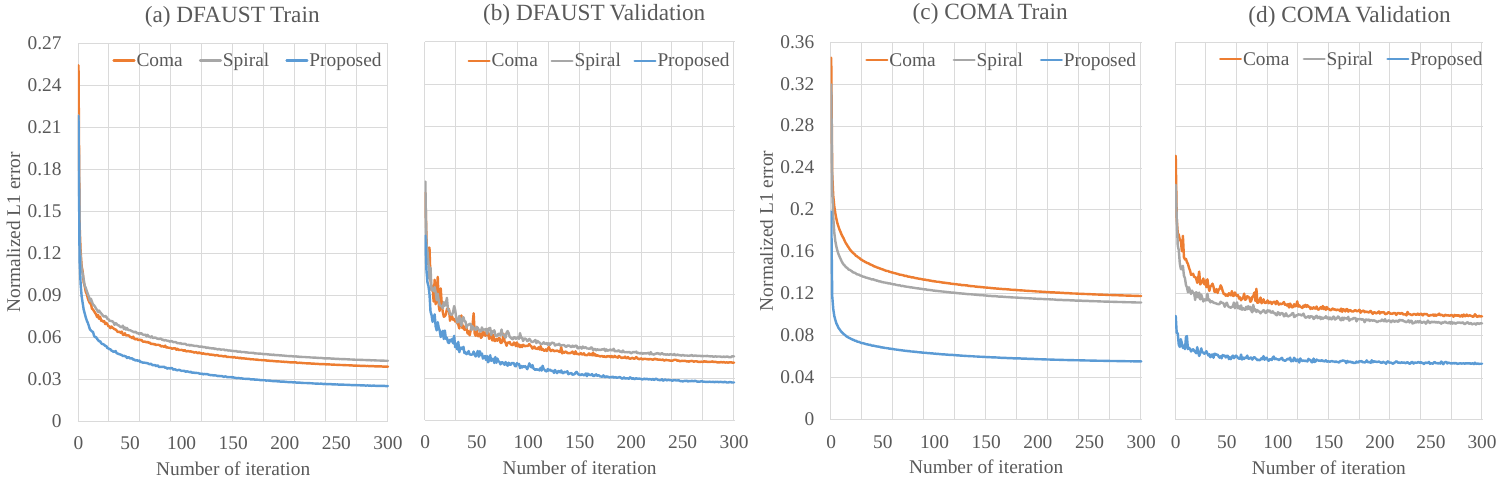}
\caption{Reconstruction errors on COMA and DFAUST datasets when the latent size $d=32$.}
\label{fig:convergence}
\end{figure*}

Figure \ref{fig:convergence} shows the validation reconstruction errors for the methods of COMA, Spiral, and our LSA-3DMM with the latent size of 32. The results of COMA and Spiral are rather close to each other: COMA performs slightly better on DFAUST dataset and Spiral performs slightly better on COMA dataset. LSA-3DMM converges much faster than COMA and Spiral. As shown in Figure \ref{fig:convergence}b, after training for only 5 iterations, LSA-3DMM achieves smaller reconstruction error than COMA and Spiral training for 300 iterations.

\begin{figure*}[tb!]
    \centering
    \includegraphics[width=0.92\textwidth]{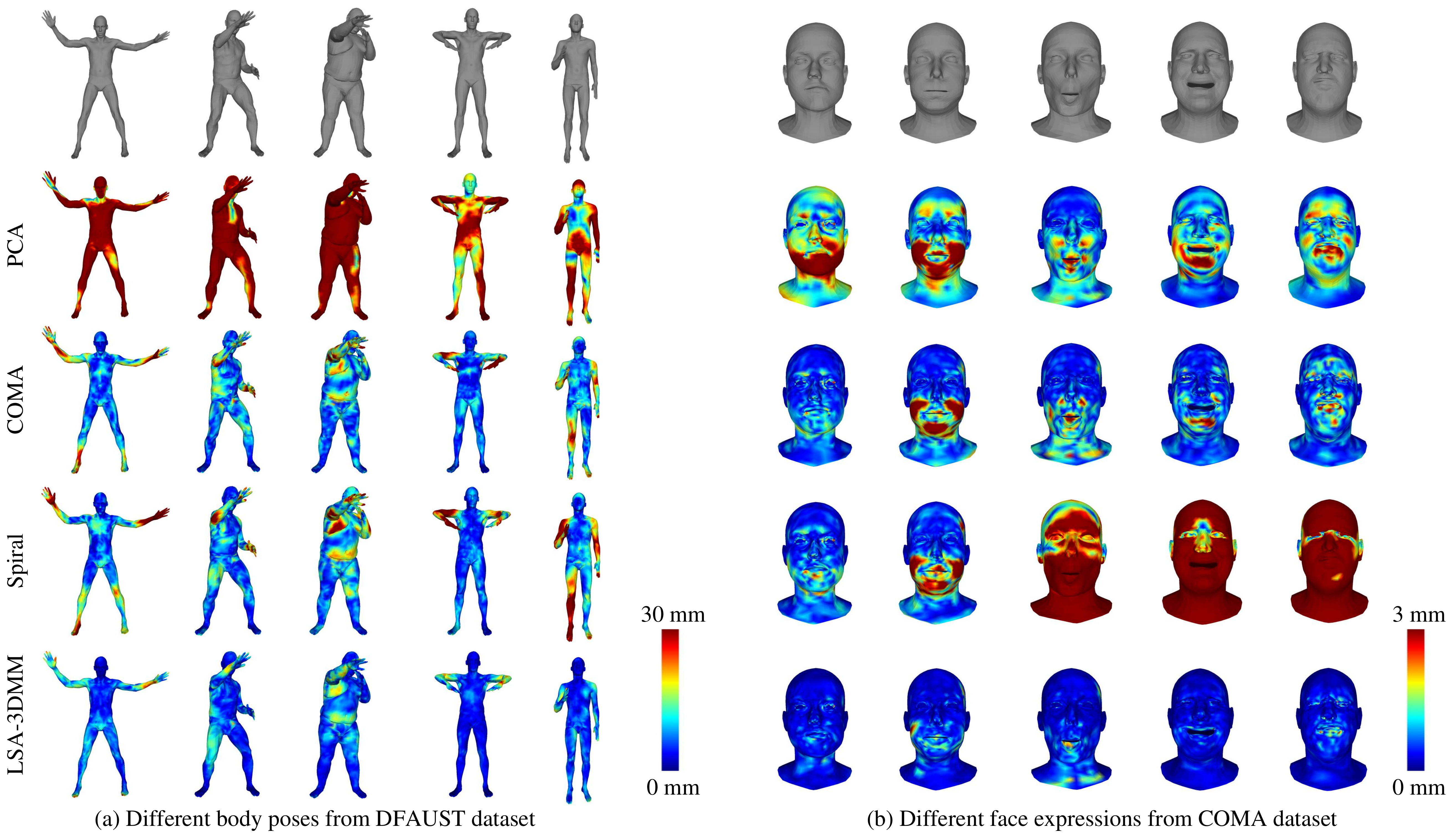}
\caption{Colormaps of per vertex Euclidean error of the reconstructions produced by PCA, COMA, Spiral, and LSA-3DMM. Top row is the ground truth meshes from test sets. Latent size $d=32$.}
\label{fig:results}
\end{figure*}

In Figure \ref{fig:results}, we qualitatively compare the reconstruction errors of some examples from the test sets of DFAUST and COMA datasets with the latent size $d=32$. Our LSA-3DMM achieves smaller reconstruction errors for each case from the test sets. It is clearly visible that PCA has the largest reconstruction errors on DFAUST dataset (Figure \ref{fig:results}a) and Spiral has the largest reconstruction errors on COMA dataset (Figure \ref{fig:results}b), which is consistent with the data shown in Figure \ref{fig:evaluation}. For Spiral, even though anisotropic filters are used compared to COMA that uses isotropic filters, the improvement is very limited since the manually designed neighboring order cannot capture the local structure very well. In contrast, our learnable weighting matrices can soft-permute each node's neighbors that cooperate well with the shared anisotropic filters to extract local features of 3D shapes. As a result, LSA-3DMM performs better for 3D shape representation learning.

\subsection{Parameter reduction for LSA-3DMM}
Our model needs to learn a weighting matrix for each node of the template. When the latent size is $d=32$, the parameter numbers of our model and PCA model are 1,867K and 482K, respectively. However, our model achieves much better reconstruction accuracy compared to PCA: $0.117mm$ vs. $0.21mm$. The proposed model is the choice for situations that require low reconstruction errors and do not strictly limit the model size. For example, in the application of monocular 3D face reconstruction \cite{gao2020semi} that uses a template with 37,202 vertices, the number of extra parameters (around 3M) of the proposed method is still relatively small compared to the backbone ResNet50 (around 23M).

We also provide a method to reduce the number of our model's parameters for situations where the model size should be limited. When the number of nodes is large, it is not necessary to learn the weighting matrix for each node since the geometric shapes of many nodes are similar to each other. We assume that the weighting matrices for all the nodes fall into a small subspace. We apply a matrix factorization technique in LSA-Conv. For meshes with $N$ nodes, we denote $\mP \in\R^{N\times K \times K}$ for all the weighting matrices that can be factorized as follows,
\begin{equation}
\mP = \mV\mP_{b},
\end{equation}
where $\mP_{b}\in\R^{B\times K\times K}$ is the $B$-dimensional subspace's bases of weighting matrices, $\mV\in\R^{N\times B}$ is the weights correspond to the $N$ nodes ($B\ll N$ e.g. $N=5023, B=8$). Instead of learning $N$ weighting matrices directly, we learn a small number of weighting matrix bases and each node's corresponding weight. This can largely reduce the number of parameters decided by the choice of $B$.

\begin{table}[tb!]
\begin{center}
\begin{small}
\begin{sc}
\begin{tabular}{l|cr|cr}
\toprule
 & \multicolumn{2}{c|}{\centering\bf DFAUST} & \multicolumn{2}{c}{\centering\bf COMA} \\
 & L2(mm) & parm \# & L2(mm) & parm \#\\
\midrule
    PCA & 9.977 & 661K & 0.210 & 482K \\
\midrule
  &\multicolumn{4}{c}{same architecture} \\
   COMA & 5.238 & 361K & 0.248 & 303K\\
   Spiral & 5.258 & 446K & 0.227 & 414K \\
   \bf LSA-3DMM & \bf 3.492 & 2,478K & \bf 0.117 & 1,867K \\
\midrule
  &\multicolumn{4}{c}{around same \# of parm} \\
  COMA~(v2) & 5.110 & 658K & 0.198 & 532K\\  
  Spiral~(v2) & 4.667 & 647K & 0.193 & 533K \\  
  \bf LSA~(small) & \bf 4.544 & 644K & \bf 0.179 & 532K \\
\bottomrule
\end{tabular}
\end{sc}
\end{small}
\caption{Comparison of reconstruction errors for the models of PCA, COMA, Spiral, and our LSA-3DMM with latent size $d=32$. \emph{LSA (small)} is the model of LSA-3DMM with parameter reduction, where the dimension of weighting matrix subspace $B=8$. COMA (v2) and Spiral (v2) are the models with increasing channel size to have around the same parameter number with \emph{LSA (small)}.}
\label{tb:results}
\end{center}
\end{table}


Table \ref{tb:results} shows the comparison of reconstruction errors for models with different parameter numbers on DFAUST and COMA datasets. When using the same architecture (Figure \ref{fig:architecture}), LSA-3DMM has the smallest reconstruction errors. We use the technique of matrix factorization to reduce the parameter number of our LSA-3DMM, denoted as \emph{LSA (small)} where the subspace dimension is set to $B=8$. For a fair comparison, we increase the channel sizes of COMA and Spiral so that the models have around the same number of parameters, denoted as \emph{COMA (v2)} and \emph{Spiral (v2)}. The channel sizes in \emph{COMA (v2)} are [64, 96, 112, 128, 128, 112, 96, 96, 64]; the channel sizes in \emph{Spiral (v2)} are [32, 64, 64, 128, 128, 110, 64, 64, 32]; while the channel sizes in \emph{LSA (small)} are [16, 32, 64, 128, 128, 64, 32, 32, 16]. In Table \ref{tb:results}, \emph{LSA (small)} achieves the best results even we use only 8 basis weighting matrices, much smaller channel size, and smaller overall parameter number. Importantly, we can balance the tradeoff between the model's reconstruction accuracy and model size by adjusting the subspace dimension.

\subsection{Ablation study}

The initial order of node's neighbors, i.e., $\{\vx_{i, 0}, \vx_{i, 1}, \ldots$, $\vx_{i, |\bm{\mathcal{N}}_i|-1}\}$, is random. In order to evaluate the robustness and effectiveness of LSA-Conv, we reshuffle the order of each node's neighbors by creating a randomized index list from 0 to $|\bm{\mathcal{N}}_i|-1$. Then we re-train the model and evaluate on the reshuffled neighbors. As shown in Table \ref{tb:random}, in both DFAUST and COMA datasets, the reconstruction errors produced by the model with neighboring reshuffle are very close to the baseline. Thus, the learnable weighting matrix in our LSA-Conv for each node is robust to the initial neighboring order. LSA-Conv is easy to be implemented without involving into any manual design for the local coordinate systems and archives remarkable results on 3D shapes.

\begin{table}[tb!]
\begin{center}
\begin{small}
\begin{sc}
\begin{tabular}{lccr}
\toprule
 & DFAUST (mm) & COMA (mm) \\
\midrule
\bf Our baseline & \bf3.492 & \bf0.117\\
Reshuffle neighbors & 3.527 & \bf0.117\\
Random init & 4.488 & 0.137\\
w/o Weighting matrix & 5.511 & 0.212\\
\bottomrule
\end{tabular}
\end{sc}
\end{small}
\caption{Ablation tests of LSA-3DMM with the latent size $d=32$. ``\emph{Reshuffle neighbors}'' means we reshuffle the order of each node's neighbors randomly. ``\emph{Random init}'' means we initialize the weighting matrices randomly from uniform distribution.}
\label{tb:random}
\end{center}
\end{table}

Table \ref{tb:random} also shows the impact of initialization methods for the weighting matrices. In baseline and ``\emph{Random init}'', we initialize the weighting matrices with identity matrix and randomly from uniform distribution, respectively. Random initialization for the weighting matrices degrades the model's performance. This is because random initialization neutralizes each node's neighbors at the beginning of training and make each node's neighbors indistinguishable, resulting in difficulty of extracting local features of 3D shapes.

We further evaluate the importance of the local structure-aware weighting matrices. Without weighting matrix, reconstruction errors of the two datasets increase. For COMA dataset, when without weighing matrix, the reconstruction error (0.212) is smaller than Spiral (0.227), meaning the predefined spiral order in SpiralNet may not be useful.

\begin{figure}[tb!]
    \centering
    \includegraphics[width=0.48\textwidth]{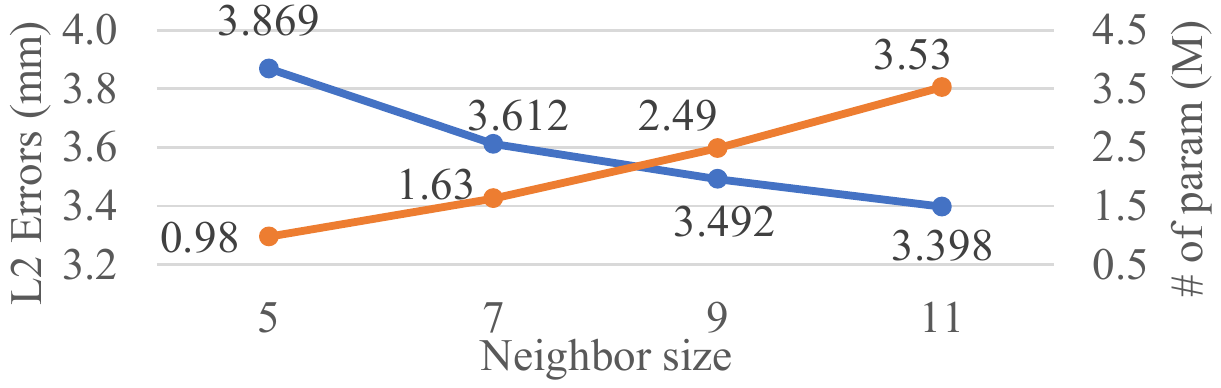}
\caption{Comparison of different neighbor sizes of LSA-3DMM on DFAUST for latent size $d=32$. Blue line denotes reconstruction error and orange curve represents model's parameter size.}
\label{fig:neighbor}
\end{figure}

In our experiments, we set the neighbor size $K=9$. The effect of the neighbor size on LSA-3DMM is shown in Figure \ref{fig:neighbor}. When we increase the neighbor size, the reconstruction error decreases and model size increases. Compared to Table \ref{tb:results}, even when the neighbor size $K=5$, the reconstruction error (3.869) are still better than PCA (9.977), COMA (5.238), and Spiral (5.258). In practice, we can choose a neighbor size to balance the tradeoff between the reconstruction accuracy and model size.

\subsection{Further applications}

\begin{figure}[tb!]
    \centering
    \includegraphics[width=0.48\textwidth]{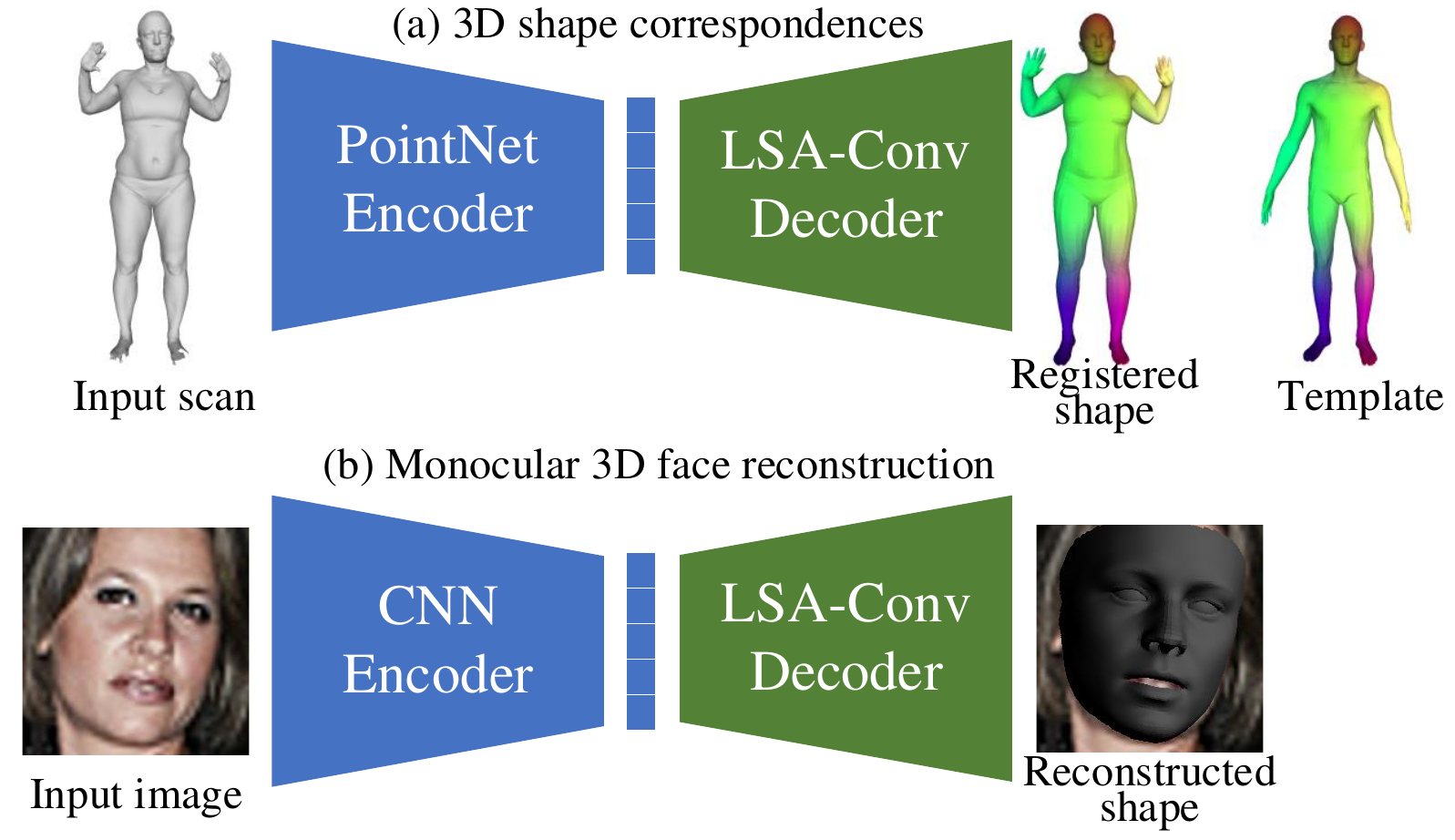}
\caption{Applications of LSA-Conv. (a) 3D shape correspondences. (b) Monocular 3D face reconstruction. Both apply encoder-decoder architecture and use LSA-Conv as a basic building block in the decoders.}
\label{fig:application}
\end{figure}

To further verify the effectiveness of our LSA-Conv, we test it on two applications: 3D shape correspondences and monocular 3D face reconstruction, shown in Figure \ref{fig:application}.

\begin{figure}[tb!]
    \centering
    \includegraphics[width=0.48\textwidth]{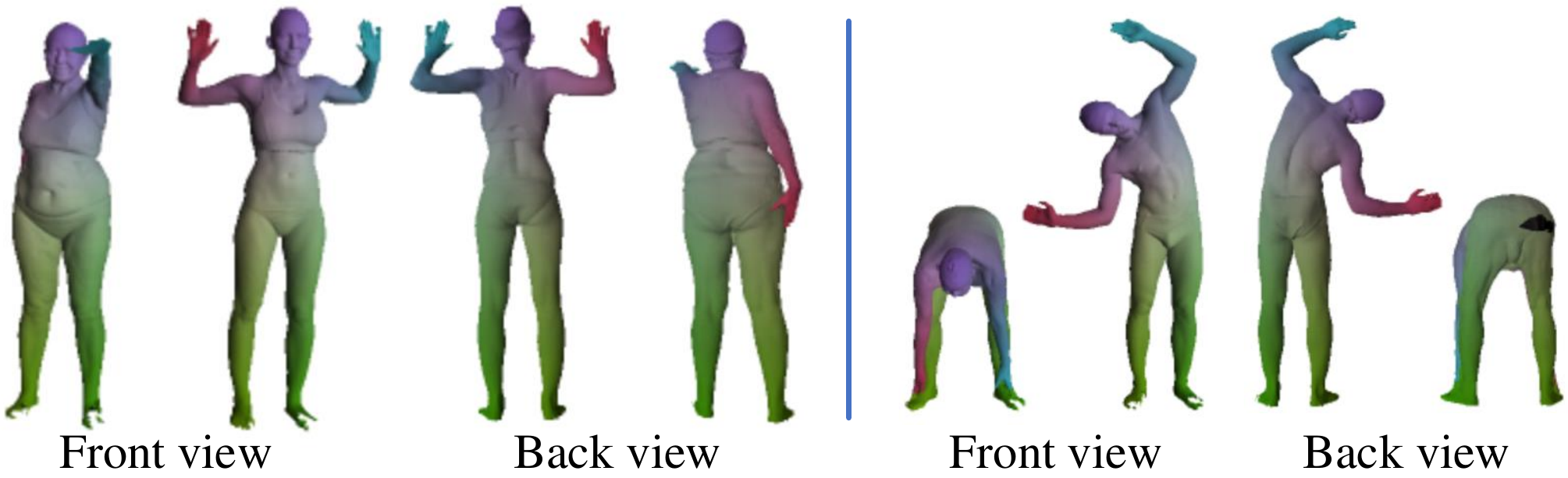}
\caption{Examples of 3D shape correspondences.}
\label{fig:registerExample}
\end{figure}

For 3D shape correspondences, we follow the pipeline of \citet{Groueix_2018_ECCV} where PointNet \cite{Qi_2017_CVPR} is used as the encoder. We only replace the shape deformation decoder with our LSA-Conv based decoder, which has channel sizes of [256, 128, 64, 64, 32, 3]. We train the network on the same synthetic data created by \citet{Groueix_2018_ECCV}, which has $2.3 \cdot 10^5$ human meshes with a large variety of realistic poses and body shapes. We evaluate our method on FAUST \cite{Bogo:CVPR:2014} that has online challenges \footnote{\url{http://faust.is.tue.mpg.de/challenge/Inter-subject_challenge}}. The error is the average Euclidean distance between the estimated projection and the ground-truth projection. For the FAUST-inter dataset, we achieve 2.501cm: \textbf{an improvement of 15\% over} \citet{Groueix_2018_ECCV}: 2.878cm. Two examples of 3D shape correspondences are presented in Figure \ref{fig:registerExample}.

\begin{figure}[tb!]
    \centering
    \includegraphics[width=0.48\textwidth]{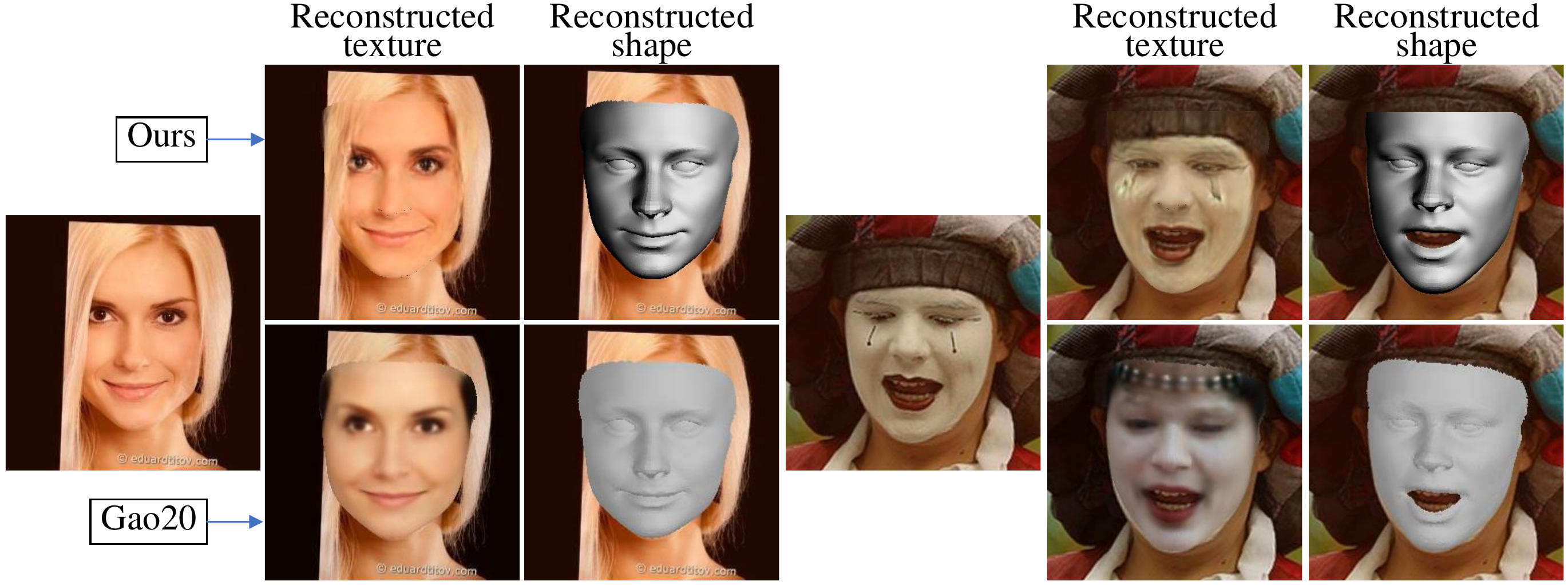}
\caption{Examples of monocular 3D face reconstruction.}
\label{fig:3dface}
\end{figure}

For monocular 3D face reconstruction, we follow the pipeline of \citet{gao2020semi}. The whole network architecture is presented in Figure 2 of \citet{gao2020semi}. ResNet-50 \cite{He_2016_CVPR} is used as the encoder. We only replace the COMA based decoder with our LSA-Conv based decoder that has channel sizes of [256, 128, 64, 32, 16, 3]. The number of vertices is 37202. We train our model on the same datasets with \citet{gao2020semi}: hybrid batches of unlabeled face images from CelebA dataset \cite{liu2015faceattributes} and labeled face images from 300W-LP dataset \cite{Zhu_2016_CVPR}. We evaluate our model qualitatively on the AFLW2000-3D dataset \cite{Zhu_2016_CVPR}. as shown in Figure \ref{fig:3dface}.

\section{Conclusion}
In this work, we propose a convolutional operation for 3D mesh representation learning and demonstrate its performance on 3D shape generation tasks, including two applications: 3D shape correspondences and monocular 3D face reconstruction. We use learnable weighting matrices to soft-permute each node's neighbors and apply shared anisotropic filters across all the nodes. Compared to previous methods that either use isotropic filters (e.g., ChebNet and GCN) or use anisotropic filters with predefined local coordinate systems (e.g., SpiralNet and MoNet), LSA-Conv is able to extract local features depending on the geometric shape of each node and has much higher representation power. Using the same architecture of convolutional mesh autoencoder, our model achieves significant improvement in 3D shape reconstruction accuracy compared to state-of-the-art methods.


\section{Acknowledgements}
This work was supported by China Major State Research Development Program (2020AAA0107600), the National Natural Science Foundation of China (U19B2035, 61831105, 61901259, 61972250), and China Postdoctoral Science Foundation (BX2019208).

\section{Ethics statement}
This paper aims to advance the technology of 3D information processing, which has wide applications in virtual reality, entertainment, sports, architecture, etc. The more accurate and efficient 3D shape representation can help people more conveniently and cost-effectively record the physical world while in the mean time, the privacy of individuals may be put at risk. Hence we shall take additional measures to protect privacy along the development of such technology.

\bibliography{mesh21}
\bibliographystyle{aaai21}

\end{document}

%% file: math_commands.tex
\usepackage{amssymb, amsmath,amsfonts,amsthm,bm} 

\def\eqref#1{equation~\ref{#1}}

\def\1{\bm{1}}

\def\vb{{\bm{b}}}

\def\vx{{\bm{x}}}
\def\vy{{\bm{y}}}

\def\mI{{\bm{I}}}

\def\mP{{\bm{P}}}

\def\mV{{\bm{V}}}
\def\mW{{\bm{W}}}
\def\mX{{\bm{X}}}

\DeclareMathAlphabet{\mathsfit}{\encodingdefault}{\sfdefault}{m}{sl}
\SetMathAlphabet{\mathsfit}{bold}{\encodingdefault}{\sfdefault}{bx}{n}

\newcommand{\R}{\mathbb{R}}

